\pdfoutput=1

\documentclass[11pt]{article}

\usepackage[]{ACL2023}

\usepackage{times}
\usepackage{latexsym}

\usepackage[T1]{fontenc}

\usepackage[utf8]{inputenc}

\usepackage{microtype}

\usepackage{inconsolata}
\usepackage{graphicx} 
\usepackage{bmpsize}
\usepackage{booktabs} 
\usepackage{CJK}
\usepackage{amsmath}
\usepackage{multirow}
\usepackage{subfigure} 
\usepackage{makecell}
\usepackage{xspace}
\usepackage{colortbl}
\usepackage{xcolor}
\usepackage{pifont}
\usepackage[misc]{ifsym}
\usepackage{balance} 
\usepackage{makecell}
\usepackage{color}
\usepackage{bbding}
\usepackage{hyperref}

\title{SSP: Self-Supervised Post-training for Conversational Search}

\author{Quan Tu$^{1}$\thanks{\ \ Equal Contribution.}\ \ , Shen Gao$^{2}$\footnotemark[1]\ \ , Xiaolong Wu$^4$, Zhao Cao$^4$, \textbf{Ji-Rong Wen}$^{1,3}$, \textbf{Rui Yan}$^{1,3}$\thanks{\ \ Corresponding author: Rui Yan (ruiyan@ruc.edu.cn).} \\
$^1$Gaoling School of Artificial Intelligence, Renmin University of China \\
$^2$School of Computer Science and Technology, Shandong University \\
$^3$Engineering Research Center of Next-Generation Intelligent \\Search and Recommendation, Ministry of Education \\
$^4$Huawei Poisson Lab \\
{\tt $^{1}$\{quantu,jrwen,ruiyan\}@ruc.edu.cn}, {\tt $^{2}$shengao@pku.edu.cn} \\
{\tt $^{4}$\{wuxiaolong19, caozhao1\}@huawei.com} \\
}

\newcommand{\ignore}[1]{}

\newcommand{\fullmodel}{\textbf{S}elf-\textbf{S}upervised \textbf{P}ost-training\xspace}
\newcommand{\model}{SSP\xspace}

\begin{document}
\maketitle
\begin{abstract}
Conversational search has been regarded as the next-generation search paradigm.
Constrained by data scarcity, most existing methods distill the well-trained ad-hoc retriever to the conversational retriever.
However, these methods, which usually initialize parameters by query reformulation to discover contextualized dependency, have trouble in understanding the dialogue structure information and struggle with contextual semantic vanishing.
In this paper, we propose \fullmodel (\model) which is a new post-training paradigm with three self-supervised tasks to efficiently initialize the conversational search model to enhance the dialogue structure and contextual semantic understanding.
Furthermore, the \model can be plugged into most of the existing conversational models to boost their performance.
To verify the effectiveness of our proposed method, we apply the conversational encoder post-trained by \model on the conversational search task using two benchmark datasets: CAsT-19 and CAsT-20.
Extensive experiments that our \model can boost the performance of several existing conversational search methods. Our source code is available at \url{https://github.com/morecry/SSP}.
\end{abstract}




\section{Introduction}

The past years have witnessed the fast progress of the ad-hoc search\cite{dai2020context,dai2018convolutional,fujiwara2013efficient, Gao2019ProductAware}. 
However, when it confronts more complicated information needs, the traditional ad-hoc search seems to be less competent. 
Recently, researchers proposes conversational search which is the combination of the search engine and the conversational assistant~\cite{radlinski2017theoretical,zhang2018towards,kiesel2021meta,trippas2020towards, 10.1145/3477495.3531830}. Different from the keyword-based query in the ad-hoc search, multi-turn natural language utterance is the main interactive form in the conversational search.
This yields the challenge of developing the conversational search system that existing ad-hoc retrievers and datasets cannot be directly used to derive the conversational query understanding module.

\begin{figure}[t!]
  \centering
  \includegraphics[width=0.9\linewidth]{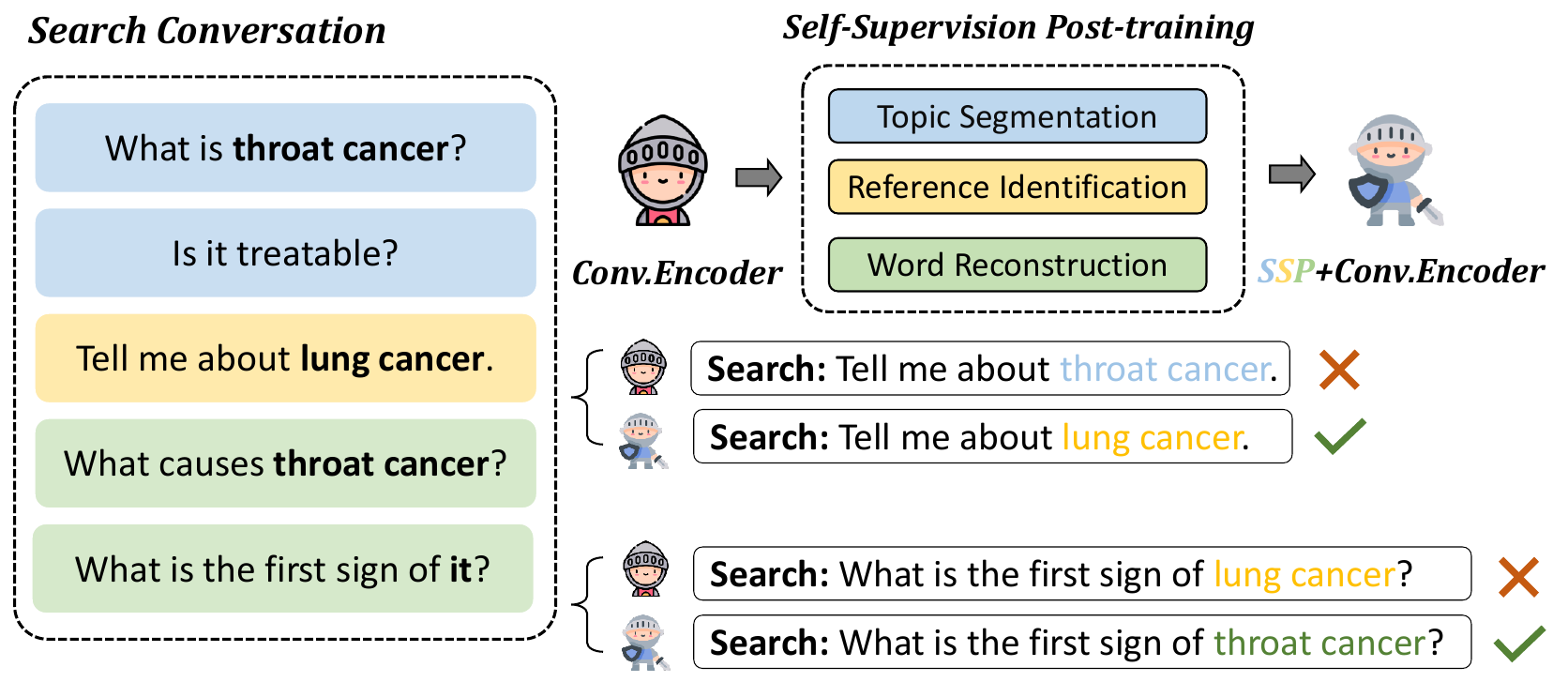}
  \caption{
  Example of modeling the conversational structure in conversational search.
  The model should capture the structure including the topic has been shifted at the 3rd utterance and the last utterance has coreference with the previous utterance. This information can help the model understand the search intent of users accurately.
  }
  \label{example}
\end{figure}


In the beginning, researchers reformulate a conversational query to a de-contextual query, which is used to perform ad-hoc retrieval~\cite{lin2020query,mele2021adaptive,lin2021multi}.
Recently the conversational dense retrieval model~\cite{lin-etal-2021-contextualized,mao2022curriculum} is presented to directly encode the whole multi-turn conversational context as a vector representation and conduct matching with the candidate document representations. 
Since the real-world conversational search corpus is hard to collect, 
a warm-up step is additionally employed to initialize the conversational representation ability~\cite{yu2021few,dai2022dialog}.
These conversational dense retrieval methods have achieved significantly better performance than the query reformulation methods and have been widely adopted in research of conversational search~\cite{yu2021few,dai2022dialog}.
However, these warm-up methods just use the same training objective on a large dataset from other domains to initialize the parameters of the conversational encoder, which can hardly capture the structure information of the conversation which is essential for understanding the user's search intent accurately.


In this paper, we propose \fullmodel (\model) for the conversational search task as shown in Figure~\ref{example}.
In \model, we replace the commonly used warm-up step with a new post-training paradigm which contains three novel self-supervised tasks to learn how to capture the structure information and keep contextual semantics. 
To be more specific, the first self-supervised task is \textit{topic segmentation}, which learns to decompose the dialogue structure into several segments based on the topic. 
To tackle the coreference problem which is a ubiquitous problem of multi-turn conversation modeling, we propose the \textit{coreference identification} task which helps the model identify the most possible referred terms in the context and simplifies the intricate dialogue structure.
Since understanding and remembering the semantic information in the conversational context is vital for conversational context modeling, we propose the \textit{word reconstruction} task which prevents contextual semantic vanishing.
To demonstrate the effectiveness of \model, we first equip several existing conversational search methods with \model and conduct experiments on two benchmark datasets: CAsT-19~\cite{dalton2020cast} and CAsT-20~\cite{dalton2021cast}.
Experimental results demonstrate that the \model outperforms all the strong baselines on $2$ datasets. 

\noindent To sum up, our contributions can be summarized as follows:

$\bullet$ We propose a general and extensible post-training framework to better initialize the conversational context encoder in the existing conversational search models.

$\bullet$ We propose three specific self-supervised tasks which help the model to capture the conversational structure information and prevent the contextual semantics from vanishing.

$\bullet$ Experiments show that our \model can boost the performance of strong conversational search methods on two benchmark datasets and achieves state-of-the-art performance.

\section{Related Work}
Conversational search has become a hop research topic in recent years.
TREC Conversational Assistant Track (CAsT) competition~\cite{dietz2017trec}, which holds the benchmark largely promotes the progress of conversational search. 
In the beginning, researchers simply view conversational search as the query reformulation problem.
They suppose that if a context-dependent query could be rewritten to a de-contextualized query based on historical queries, then it directly uses the well-trained ad-hoc retriever to obtain retrieval results. 
Transformer++~\cite{vakulenko2021question} fine-tunes the GPT-2 on query reformulation dataset CANARD~\cite{elgohary2019can} to rewrite query.
QueryRewriter~\cite{yu2020few} exploits large amounts of ad-hoc search sessions to build a weak-supervision query reformulation data generator, then these automatically generated data is used to fine-tune the language model.  
However, these methods underestimate the value of context, which contains various latent search intentions and topic information. 

\begin{figure}[ht]
\vspace{
-1mm}
  \centering
      \includegraphics[width=0.9\linewidth]{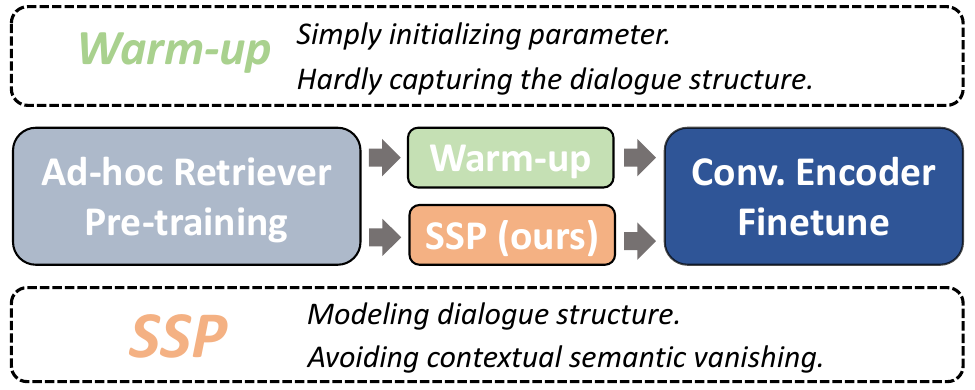}
    \caption{The comparison between the training procedure of conversational search with warm-up and the \model paradigm.}  
  \vspace{-3mm}
  \label{framework comparison}
\end{figure}

After that, the conversational dense retriever is proposed. 
It straightly encodes full conversation whose last query denotes the user's real search intention to dense representation. 
ConvDR~\cite{yu2021few} forces the contextual representation to mimic the reformulation query representation based on the teacher-student framework, which slightly deals with the conversational search data scarcity problem. 
Further, COTED~\cite{mao2022curriculum} points out that not all queries in context are useful and devises a curriculum denoising method to inhibit the influence of unnecessary contextual queries. 
These dense methods additionally perform the warm-up on the other domain dataset to initialize the parameters based on their own objective.
However, their warm-up ignore the conversation structure information, which is crucial for capturing the relationship between utterances and understanding the search intention of the user. In this respect, we devise a novel \fullmodel(\model) to replace the warm-up as Figure~\ref{framework comparison}. 

\section{Problem formulation}
\label{sec:formulation}

We assume that there is a multi-turn search conversation $Q=\{q_1, q_2, \dots, q_{n}\}$, where $q_i=\{x_{i,1}, x_{i,2}, \dots, x_{i,l_i}\}$ represents the $i$-th question in the conversation and $x_{i,j}$ is the $j$-th token in $q_i$.
The last query $q_n$ is the user's real search intention. 
We insert special tokens $\mathtt{[CLS]}$ and $\mathtt{[SEP]}$ in $Q$ yielding $\{\mathtt{CLS}, q_1, \mathtt{[SEP]}, q_2, \allowbreak \mathtt{[SEP]}, \dots, \mathtt{[SEP]}, q_{n}\}$ as the model input, where $\mathtt{[CLS]}$ is the start token and $\mathtt{[SEP]}$ is the separation token to split each query. 
After the concatenation of all queries is sent into the conversational encoder (a transformer-based architecture model), we obtain the last layer's output hidden state $E$. $E_{\mathtt{[CLS]}}$ and $E_{\mathtt{[SEP]}}$ are the corresponding representations of $\mathtt{[CLS]}$ and $\mathtt{[SEP]}$ and will be used in self-supervised tasks. 
Our goal is to learn a better contextual representation $E_{\mathtt{[CLS]}}$ in order to accurately retrieve documents in corpus for the last query $q_n$.

\section{Self-Supervised Post-training}

\begin{figure*}[ht]
  \centering
      \includegraphics[width=0.95\linewidth]{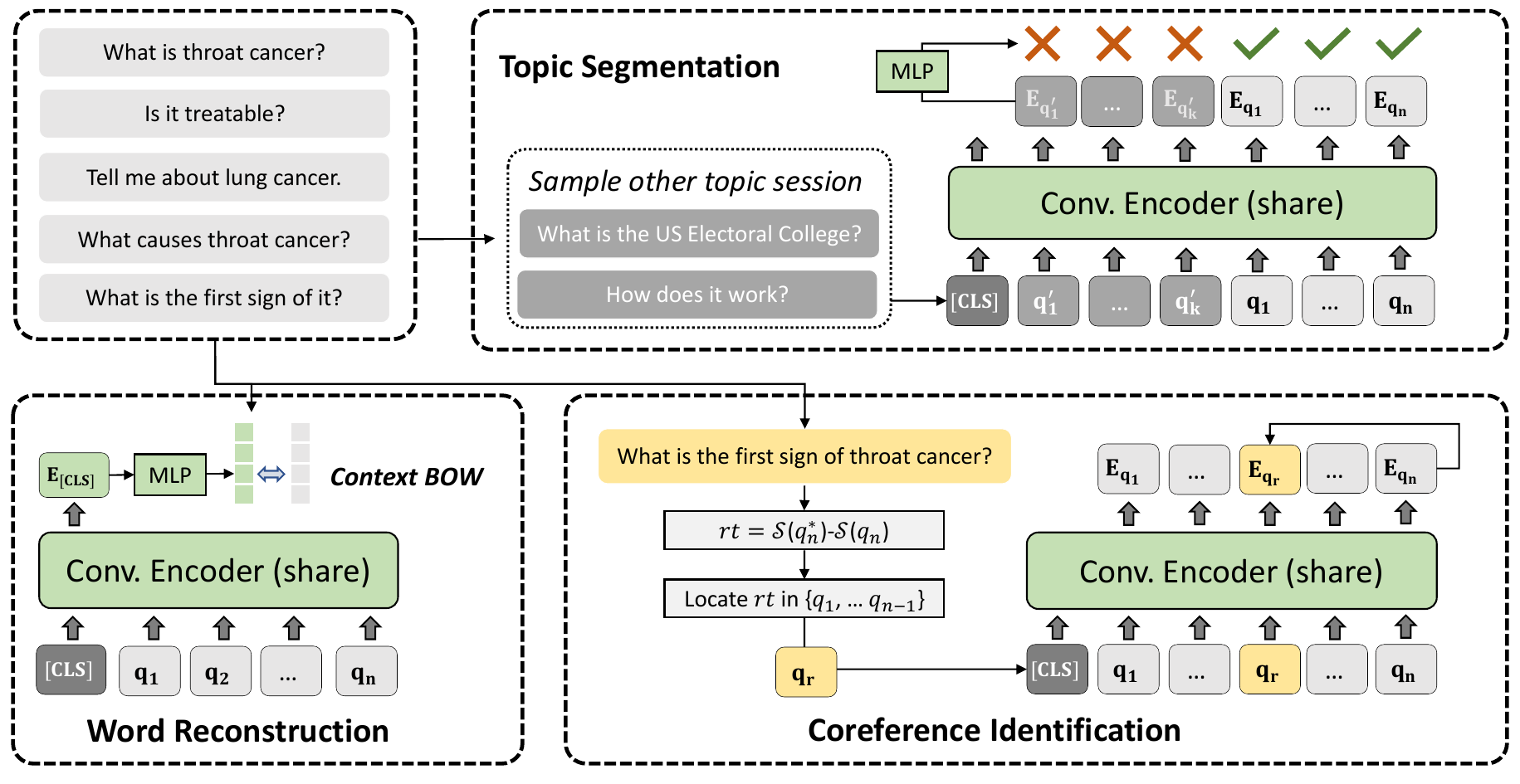}
  \caption{Overview of \model. It consists of three self-supervised tasks to conduct post-training of conversational encoder: (1) \textit{Topic Segmentation} predicts which utterances are the randomly sampled perturbation utterances from other conversation sessions; (2) \textit{Coreference Identification} predicts which utterance in the conversational context is related to the last utterance; (3) \textit{Word Reconstruction} uses the conversational context vector representation to reconstruct the Bag-of-Word vector of conversational context.}
  \label{model}
\end{figure*}

\subsection{Overview}

In this section, we propose our \fullmodel, abbreviated as \model. 
An overview of \model is shown in Figure~\ref{model}, which consists of three self-supervised tasks:

$\bullet$ \textit{Topic Segmentation Task} aims to find the topic-shifting point in the utterances. It helps the model to capture the topic structure in the conversational context.

$\bullet$ \textit{Coreference Identification Task} aims to identify the correlation structure between two referred utterances, which helps the conversational encoder to understand the coreference relationship and produce better query representation.

$\bullet$ \textit{Word Reconstruction Task} aims to reconstruct the bag-of-word (BOW) vector of the conversational context using the conversational vector representation. It helps the model avoid the contextual semantic vanishing during conversation encoding.

After jointly training the conversational encoder using these three self-supervised tasks, we fine-tune the encoder to the conversational search downstream task using the existing conversational search methods.


    
\subsection{Topic Segmentation Task} 
\label{sec:topic-seg}

When the user interacts with the conversational search system, the focused topic may vary from time to time.
Taking the example in Figure~\ref{example}, the search intention of the user changes according to the retrieval results of previous turns.
This causes the topic of the conversation to shift.
Since the conversation topic may shift in every utterance, to fully understand a user query, the conversational system should know what is the current topic of this query and view the utterances of the current topic as a more salient context.
If the conversational encoder cannot identify the topic boundary of the current topic, it may focus on unrelated utterances and incorporate noise information into the query representation.


 Thus we propose the topic segmentation task to identify the topic boundary of the conversation, which can facilitate the model to focus on more related context when encoding the query.
We first randomly sample a noise conversational session with several utterances from the training corpus and then concatenate this sampled noise session at the beginning of the raw conversational context.
Given the raw search conversation $Q=\{q_1, q_2, \dots, q_{n}\}$ and the noisy conversation $Q^{\prime}=\{q^{\prime}_1, q^{\prime}_2, \dots, q^{\prime}_{m}\}$, we truncate the first $k$ queries of $Q^{\prime}$ where $k$ is sampled based on reciprocal probability distribution $p$, which avoids the distortion of the raw context from the abundant long noisy sessions,
\begin{equation*}
    p_{k} = {\frac{1}{k}} / {\sum_{i=1}^{m}\frac{1}{i}}, {k= 1, 2, \dots, m}. 
\end{equation*}
After concatenating the sampled noise session before the raw context and separating each query by $\mathtt{[SEP]}$, we obtain the perturbed conversation $\check{Q}=\{\mathtt{[CLS]}, q^{\prime}_1, \mathtt{[SEP]}, \allowbreak \dots, q^{\prime}_{k}, \mathtt{[SEP]}, q_1, \mathtt{[SEP]}, \dots, q_{n}\}$ and the ground truth topic label $y^{t}=\{1, \dots, 1, 0, \dots, 0\}$, where the queries from the external conversation are labelled as $1$ and the ones from the raw conversation are labelled as $0$. 

Next, we use the perturbed conversation $\check{Q}$ as input to the conversational encoder, and obtain the vector representation  $\check{E} = \{E_{\mathtt{[CLS]}}, E^{\prime}_{1}, E_{\mathtt{[SEP]}}, \dots,\allowbreak  E^{\prime}_{k}, E_{\mathtt{[SEP]}}, E_{1}, \allowbreak E_{\mathtt{[SEP]}}, \dots, E_{n}\}$ of the perturbed conversation $\check{Q}$.
Finally, $E_{\mathtt{[SEP]}}$ is sent to the topic predictor (a linear layer) to decide whether an utterance is from the sampled noise conversation $Q^{\prime}$ or not. 
The binary cross entropy is used to compute topic segmentation loss $\mathcal{L}_{TS}$: 
\begin{equation*}
    \begin{aligned}
    p(y^{t}_i=1|\check{Q}) &= \text{Sigmoid}(W_{t}E_{\mathtt{[SEP]}}+b_{t}),\\
    \mathcal{L}_{TS} &= -y^{t}_i\log(p(y^{t}_i=1|\check{Q})) \\
    &- (1-y^{t}_i)(1-\log(p(y^{t}_i=1|\check{Q}))),
    \end{aligned}
\end{equation*}
where $W_t \in \mathbf{R}^{h\times1}, b_t \in \mathbf{R}$, $h$ is the hidden size of model. 

\subsection{Coreference Identification Task}

In conversational search, a common problem is the coreference, which is that the pronoun in a query usually refers to a term in its previous queries.
Most of the existing methods did not explicitly train the model to tackle this problem.
Here, we devise an auxiliary self-supervised task that trains the model to predict the referred utterance of the last utterance by the coreference relationship.
To obtain which utterance in the conversational context has the coreference relationship with the last utterance, we use the query reformulation corpus to find. 
We compare the last query in $Q$ with the reformulated query $q^{*}_{n}$ by set operations to find the reformulation terms $r$ have been omitted in $Q$:
\begin{equation*}
     r = \mathcal{S}(\text{tokenize}(q^{*}_{n})) - \mathcal{S}(\text{tokenize}(q_{n})), 
\end{equation*}
where $\mathcal{S}$ is a set operation that converts a sentence into a non-repeating word set. We can obtain the reformulation terms $r$ by calculating the difference set between two sets.
Then $r$ will be used to locate the referred query from back to front until the first query containing the $r$ is found. 
We mark the position of the referred query to the label $y^{c}=\{0,0, \dots, 1, \dots, 0\}$, whose $i$-th value is 1 only if the $i$-th query is the referred query. 
Similar to the topic segmentation task (introduced in \S~\ref{sec:topic-seg}), we send $E_{\mathtt{[SEP]}}$ into a coreference predictor to predict the referred query and use the binary cross-entropy as the loss function of this task: 
\begin{equation*}
    \begin{aligned}
    p(y^{c}_i=1|Q) &= \text{Sigmoid}(W_{c}E_{\mathtt{[SEP]}}+b_{c}), \\
    \mathcal{L}_{CI} &= -y^{c}_i\text{log}(p(y^{c}_i=1|Q)) \\
    &- (1-y^{c}_i)(1-\text{log}(p(y^{c}_i=1|Q))),
    \end{aligned}
\end{equation*}
where $W_r \in \mathbf{R}^{h\times1}, b_r \in \mathbf{R}$ are all trainable parameters. With the coreference identification task, the conversational encoder will pay more attention to the most possible referred query in context when it understands the last query. 

\subsection{Word Reconstruction Task}

The duality of a one-stage conversational retriever will encode a query to a dense vector. 
In the previous sections, we use the self-supervised tasks to focus on the utterance of the current topic and the highly related utterance with coreference.
However, other utterances may also provide useful information to understand the current search intent.
Thus, the conversational encoder should not only gather information from the related utterances but also keep the information from the whole conversational context.

To avoid the information vanishing in the final conversational vector representation, we propose to use a simple but efficient reconstruction task to help the conversational encoder to keep the overall semantic information.
In this task, we train the model to reconstruct the bag-of-words (BOW) vector of the whole conversation using the representation of $\mathtt{[CLS]}$ produced by the conversational encoder.
Specifically, all of the words appearing in the context are converted to a BoW vector $y^w$, 
\begin{equation*}
    y^w = \text{BOW}(\mathcal{S}(\text{tokenize}(Q))),
\end{equation*}
where the length of $y^w$ is the vocab size and $y^w_{i}=1$ only if the $i$-th word in vocab appears in the context otherwise $y^w_{i}=0$. We use a linear layer after the last layer of the model to process $E_{\mathtt{[CLS]}}$ and optimize the WR loss based on mean squared error, 

\begin{equation*}
    \begin{aligned}
    \hat{y}^{w} &= \text{Sigmoid}(W_{w}E_{\mathtt{[SEP]}}+b_{w}), \\
    \mathcal{L}_{WR} &= \left\|\hat{y}^{w}-y^w\right\|_2,
    \end{aligned}
\end{equation*}
where $W_w \in \mathbf{R}^{h\times|V|}, b_w \in \mathbf{R}^{|V|}$, $|V|$ is the vocab size, $\left\|\cdot\right\|$ means euclidean distance. 

\subsection{Optimization}

Inspired from the previous studies~\cite{yu2021few, mao2022curriculum}, we also employ the knowledge distillation objective in \model to accelerate the learning process.
Specifically, a pre-trained ad-hoc search encoder $\text{TEnc}$ which uses the de-contextualized query as the input and produce the vector representation.
We use $\text{TEnc}$ as the teacher model and employ a knowledge distillation loss function to train our conversational encoder to mimic the vector representation produced by the teacher encoder $\text{TEnc}$.
We formulate the knowledge distillation loss $\mathcal{L}_{KD}$ as follows:
\begin{equation*}
    \begin{aligned}
    E^{*}_{\mathtt{[CLS]}} &= \text{TEnc}(\{\mathtt{[CLS]}, q^{*}_{n}\})_{\mathtt{[CLS]}} \\
    \mathcal{L}_{KD} &= \left\|E_{\mathtt{[CLS]}}-E^{*}_{\mathtt{[CLS]}}\right\|_2. 
    \end{aligned}
\end{equation*}
where the $q^{*}_{n}$ is the manual rewritten query of $q_{n}$, $(\cdot)_{\mathtt{[CLS]}}$ means only taking the $\mathtt{[CLS]}$ representation of TEnc's last layer output. We make the representation of conversation $E_{\mathtt{[CLS]}}$ to approximate the representation of reformulation query $E^{*}_{\mathtt{[CLS]}}$ processed by \text{TEnc} to distill its powerful retrieval ability.

Finally, we combine all the training objective of each self-supervised task and optimize all the parameters in the conversational encoder:
\begin{equation*}
     \mathcal{L}_{\text{final}}= \mathcal{L}_{KD} + \alpha\mathcal{L}_{TS} + \beta\mathcal{L}_{CI} + \gamma\mathcal{L}_{WR},
\end{equation*}
where the $\mathcal{L}_{\text{final}}$ is the final training objective for \model, $\alpha, \beta$, and $\gamma$ denotes the hyper-parameter as a trade-off between the self-supervised tasks.
\section{Experimental Setting}


\subsection{Datasets}

\begin{table}[!h]
  \centering
  \small
  \caption{The statistics of test dataset for fine-tuning.}
    \begin{tabular}{l|c|c}
    \toprule
    \textbf{Statistics} & \textbf{CAsT-19} & \textbf{CAsT-20} \\
    \midrule
    \# Conversations & 50 (20) & 25 (25) \\
    \# Queries  & 479 (173) & 216 (208) \\
    \# Avg. Query Tokens & 6.1 & 6.8 \\
    \#  Avg. Queris / Conversation & 9.6 & 8.6 \\
    \midrule
    \# Documents & \multicolumn{2}{c}{38M} \\
    \bottomrule
    \end{tabular}%
  \label{tab:statistics of cast}%
\end{table}%

For fine-tuning the conversational encoder on the conversational search task, we choose two few-shot datasets to evaluate our proposed model based on K-fold cross-validation.

\textbf{CAsT-19}~\cite{dalton2020cast} is the acronym of the TREC Conversational Assistance Track (CAsT) 2019 benchmark dataset. 
It is built by human annotators who are required to mimic real dialogues under specified topics and contains frequent coreferences, abbreviations, and omissions. 
In this work, we pay attention to query de-contextualization and but only the test set provides manual oracle de-contextualized queries. 
Since the queries in TREC CAsT dataset are used in the conversational search fine-tuning phrase, it will cause the data leaking problem.
For a fair comparison, we filter the queries from TREC CAsT from QReCC. 
The statistics of the filtered QReCC dataset are shown in Table~\ref{tab:statistics of qrecc}. 

\textbf{CAsT-20}~\cite{dalton2021cast} refers to next year's TREC CAsT. 
Its most obvious modification is that the coreference could appear in the response (a summarized answer of gold passage)compared with CAsT-19, where a query only refers to its previous queries. 
Both manual response and automatic response (generated by neural rewriter~\cite{yu2020few}) are provided in CAsT-20. 
It contains 216 queries in 25 dialogues which have de-contextualized queries and most of queries have relevance judgments. 
Additionally, CAsT-20's corpus is the same as CAsT-19's. 
Detailed statistics are shown in Table~\ref{tab:statistics of cast}.

\subsection{Baselines}

Following~\cite{mao2022curriculum}, we split baselines into two categories: sparse retrieval methods and dense retrieval methods respectively. Sparse retrieval methods rewrite the contextualized query to a context-independent query and use the ad-hoc sparse retriever to obtain the results.
The dense retrieval methods use the ad-hoc dense retriever or directly encode the conversational queries via a conversational dense retriever. 
     
$\bullet$ \texttt{Raw} denotes simply using the last context-independent query in the dense or sparse retriever to retrieve the documents.

$\bullet$ \texttt{Tansformer++}~\cite{vakulenko2021question} is a query rewriting method which inherits from GPT-2~\cite{radford2019language} and fine-tunes on CANARD dataset~\cite{elgohary2019can}. Then it employs the ad-hoc retriever to search using the rewritten query.

$\bullet$ \texttt{QueryRewriter}~\cite{yu2020few} is a data augmentation method that first generates query reformulation data using large amounts of ad-hoc search sessions based on rules and self-supervised learning. Then the automatically generated data is used to train the query rewriter.

$\bullet$ \texttt{QuReTeC}~\cite{voskarides2020query} deals with the query reformulation task as a binary term classification problem. It will decide whether to add terms appearing in the dialogue history to the current turn query or not. 

$\bullet$ \texttt{ContQE}~\cite{lin-etal-2021-contextualized} employs a well-trained ad-hoc search encoder TCT-ColBERT~\cite{lin2020distilling}. It uses the mean-pooling method to get the contextual embedding and fine-tunes on pseudo-relevance labels.

$\bullet$ \texttt{ConvDR}~\cite{yu2021few} develops the few-shot learning method to train the conversational dense retriever. It takes ANCE~\cite{xiong2020approximate} as the teacher model to teach the conversational student model. Integrating the distilling loss and ranking loss, it obtains a pretty performance on the few-shot dataset.

$\bullet$ \texttt{COTED}~\cite{mao2022curriculum} further introduces the curriculum denoising to inhibit the unhelpful turns in context. An additional two-step multi-task learning improves the performance of \texttt{ConvDR}.

$\bullet$ \texttt{T5(WikiD+WebD)}~\cite{dai2022dialog} trains on two large automatically generated conversational search dataset WikiDialog(11.4M dialogues) and WebDialog(8.4M dialogues) from a T5-large encoder checkpoint. Otherwise, it further warm-ups on the QReCC dataset. Though it does not fine-tune on CAsT-19 (50 dialogues) and CAsT-20 (25 dialogues), the extremely time-consuming training procedure makes its performance up to a stable level.

\subsection{Evaluation Metrics}
Following the previous works on conversational search, we evaluate all models based on \textbf{M}ean \textbf{R}eciprocal \textbf{R}ank (MRR) and \textbf{N}ormalized \textbf{D}iscounted \textbf{C}umulative \textbf{G}ain @3 (NDCG@3).
\textbf{MRR} deems the ranking reciprocal of a positive sample as its score and counts the average of all samples. It is a simple yet effective metric for ranking tasks. 
\textbf{NDCG@3} considers the importance of positive samples based on their relevance and chooses scores of the top 3 samples to normalize. 
The statistical significance of two runs is tested using a two-tailed paired t-test and is denoted using $\dagger$ and $\ddagger$ for significance ($p \le 0.05$) and strong significance ($p \le 0.01$).

\subsection{Implementation Details}
Most settings in this work are similar to ConvDR~\cite{yu2021few}. 
We employ the ad-hoc retriever ANCE~\cite{xiong2020approximate} as the teacher module to calculate the knowledge distillation loss.
Following previous conversational search work, for CAsT-19, we concatenate the historical query and the current query as the model inputs, and we additionally take account of the historical responses for CAsT-20.
The leading words in the conversational context will be truncated if the concatenation length exceeds a maximum length, which is 256 and 512 for CAsT-19 and CAsT-20 respectively.
We implement experiments using PyTorch and Transformers library on an NVIDIA A40 GPU. 
Adam optimizer is employed with the learning rate of $2e-5$ and batch size of $64$ for CAsT-19 and $32$ for CAsT-20. 
Our model will post-train $2$ epochs and then fine-tune on the conversational search corpus. 
The self-supervised task weights $\alpha, \beta$ and $\gamma$ are set as $1e-2, 1e-3, 1e-2$ for CAsT-19 and $ 1e-1, 2e-3, 2e-2$ for CAsT-20.
We use faiss~\cite{johnson2019billion} to index the passages, whose representations are generated by ANCE and fixed. 
Following the TREC Conversational Assistance competition official evaluation setting, we use relevance scale $\le$ 2 as positive for CAsT-19 and relevance scale $\le$ 1 for CAsT-20 and obtain our result based on official evaluation scripts. 
\section{Evaluation Result}
\subsection{Overall Performance}

\begin{table}[t!]
  \centering
  \small
  \caption{Conversational search performance comparison. $\star$ denotes our implementation. $\dagger$ ($\ddagger$) indicates (strong) significant improvement over \texttt{ConvDR} with $p \le 0.05$ ($p \le 0.01$).}\label{tab:main_result}%
  \resizebox{1.0\columnwidth}{!}{
    \begin{tabular}{c|c|cc|cc}
    \toprule
    \multirow{2}{*}{Search} & \multirow{2}{*}{Method} & \multicolumn{2}{c|}{CAsT-19} & \multicolumn{2}{c}{CAsT-20} \\
    \cline{3-6}          &       & MRR & NDCG@3 & MRR & NDCG@3 \\
    \midrule
    \multirow{4}[2]{*}{Sparse} & \texttt{Raw}   & 0.322 & 0.134 & 0.160  & 0.101 \\
          & \texttt{Tansformer++}  & 0.557 & 0.267 & 0.162 & 0.100 \\
          & \texttt{QueryRewriter} & 0.581 & 0.277 & 0.250  & 0.159 \\
          & \texttt{QuReTeC} & 0.605 & 0.338 & 0.262 & 0.171 \\
    \midrule
    \multirow{12}[4]{*}{Dense} & \texttt{Raw}   & 0.420  & 0.247 & 0.234 & 0.150 \\
          & \texttt{Tansformer++} & 0.696 & 0.441 & 0.296 & 0.185 \\
          & \texttt{QueryRewriter} & 0.665 & 0.409 & 0.375 & 0.255 \\
          & \texttt{QuReTeC} & 0.709 & 0.443 & 0.430  & 0.287 \\
          & \texttt{ContQE} & -     & \textbf{0.499} & -     & 0.312 \\
          
          & \texttt{T5(WikiD+WebD)} & 0.741  & -  & 0.513  & -  \\
          & \texttt{COTED} & 0.769  & 0.478  & 0.491  & 0.342  \\ 
\cline{2-6} 
& \texttt{COTED}$^{\star}$   & 0.758  & 0.475   & 0.481  & 0.321  \\
& \texttt{COTED-\model}    & 0.760   & 0.478  & 0.501 & 0.351 \\
\cline{2-6} 
& \texttt{ConvDR} & 0.740  & 0.466  & 0.501  & 0.340  \\
& \texttt{ConvDR-\model}  & \textbf{0.780}$^{\dagger}$      & 0.480 & \textbf{0.526}$^{\ddagger}$ & \textbf{0.365}$^{\ddagger}$  \\
    \bottomrule
    \end{tabular}%
    }
\end{table}%

We compare our model with all baselines in Table~\ref{tab:main_result}.
We can find that the sparse methods generally achieve less satisfying performance than the dense conversational methods, which demonstrates the dense methods can understand the search intent of users better.
Our model performs consistently better on two datasets than other sparse and dense conversational search models with improvements of 1.4\% and 0.4\% on the CAsT-19 dataset and achieves 7.1\% and 6.7\% improvements on the CAsT-20 dataset compared with \texttt{COTED} in terms of MRR, and NDCG@3 respectively.
This demonstrates that our proposed self-supervised tasks provide a useful training signal for the conversational encoder module than the simple parameter warm-up method used in previous methods.

In Table~\ref{tab:main_result}, we find that \texttt{ContQE} outperforms \texttt{ConvDR-\model} on CAsT-19 in terms of NDCG@3. 
The possible reason is that \citet{mao2022curriculum} has illustrated that \texttt{ContQE} introduces a stronger query encoder TCT-ColBERT~\cite{lin2020distilling} and it takes multi-stage methods to train their conversational encoder.
In contrast to the complexity of the multi-stage method, our \model can boost the performance of the existing conversational search model in an end-to-end manner which is easier to train and deploy in real-world applications.
We will leave adapting this stronger encoder TCT-ColBERT into the post-training paradigm in our future work.

To verify the generalization ability of \model, we equip our proposed \fullmodel to two strong conversational search methods (\texttt{COTED} and \texttt{ConvDR}), which can provide a better conversational context encoder.
From the comparison between \texttt{COTED} and \texttt{COTED-\model}, \texttt{ConvDR} and \texttt{ConvDR-\model}, we can find that our proposed new post-train paradigm can adapt to different conversational search models and boost their performance, which demonstrate the effectiveness and generalization ability of our proposed \model.

\subsection{Ablation Study}

\begin{table}[t!]
\small
  \centering
   \caption{Comparison between ablation models.}
    \begin{tabular}{c|cc|cc}
    \toprule
    \multirow{2}{*}{Method} & \multicolumn{2}{c|}{CAsT-19} & \multicolumn{2}{c}{CAsT-20} \\
\cline{2-5}          & MRR   & NDCG@3 & MRR   & NDCG@3 \\
    \midrule
    \texttt{ConvDR-\model}  & \textbf{0.780} & \textbf{0.480 } & \textbf{0.526 } & \textbf{0.365 } \\
    \texttt{w/o. TS} & 0.753  & 0.473  & 0.513  & 0.355  \\
    \texttt{w/o. CI} & 0.749  & 0.472  & 0.515  & 0.351  \\
    \texttt{w/o. WR} & 0.757  & 0.476  & 0.512  & 0.357  \\
    \bottomrule
    \end{tabular}%
   \label{tab:ablation}%
\end{table}%

We remove each self-supervised task to analyze the effectiveness of each component, and TS is the acronym for topic segmentation, CI denotes the coreference identification and WR denotes word reconstruction.
The performance of ablation models is shown in Table~\ref{tab:ablation}, and we can find that all of the ablation models perform less promising than the best model \texttt{ConvDR-\model}, which demonstrates the preeminence of each self-supervised task in \model. 

We ablate the topic segmentation task in \texttt{ConvDR-\model w/o. TS} and observe the decline in search performance. 
The topic segmentation task helps the model identify the topic boundary in the long session and pay more attention to the utterances in the related topics 
This makes the retrieval performance raises $3.6\%$ and $2.5\%$ in terms of MRR on the CAsT-19 and CAsT-20 datasets respectively.
In the method \texttt{ConvDR-\model w/o. CI}, we remove the coreference identification self-supervised task and the performance of this ablation model dropped dramatically, which demonstrates that it plays the most important role in \model.
The experiment shows that our \texttt{ConvDR-\model} achieves 4.1\% and 1.7\% increments compared with \texttt{ConvDR-\model w/o. CI} in terms of MRR score on the CAsT-19 and CAsT-20 datasets. 
We also remove the word reconstruction task yielding \texttt{ConvDR-\model w/o. WR}, and the dropped score shows that it is effective to keep the contextual semantic in the context representation.
All of our self-supervised tasks, which provide extra supervision signals to understand dialog structure and prevent the semantic vanishing, help \texttt{ConvDR-\model} achieves the best performance according to the experimental results.

\subsection{Robustness of Topic Segmentation}

\begin{figure}[t!]
\centering
  \hspace{-3mm}
      \includegraphics[width=0.95\linewidth]{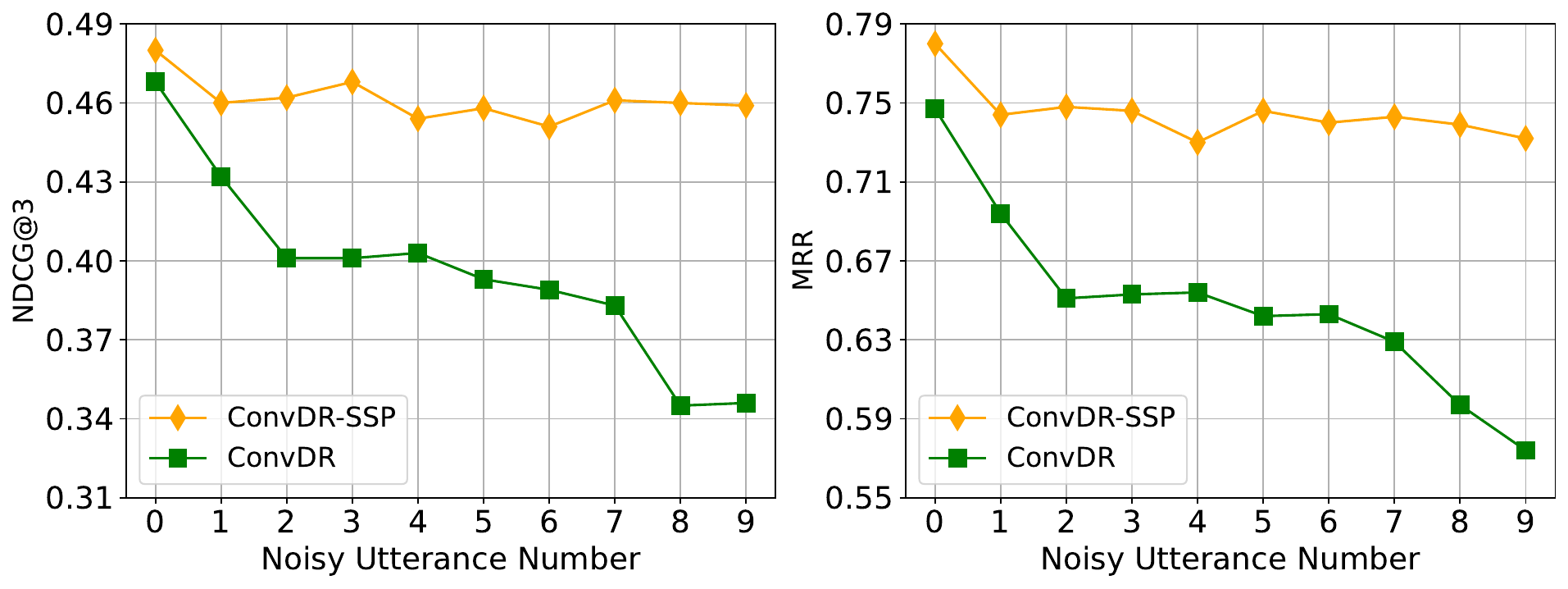}
  \caption{Robustness evaluation by adding the different numbers of off-topic utterances. We randomly sample irrelevant utterances from other search sessions and evaluate the results of \texttt{ConvDR} and \texttt{ConvDR-\model}.}
  \label{fig:add_turn}
\end{figure}

\begin{table*}[ht]
\vspace{3mm}
\small
     \caption{Retrieved examples of \texttt{ConvDR-\model} and \texttt{ConvDR}. We present historical queries, current query (underlined), manual reformulation query (\textbf{Ref}) and the first passages different methods disagree. The key information in the conversations and passages are marked in \textcolor{blue}{blue} and \textcolor{red}{red} respectively.}
    \label{tab:case} 
    \resizebox{0.99\linewidth}{!}{
    \begin{tabular}{p{0.48\linewidth}@{} |lp{0.51\linewidth}} 
    \hline
    \textbf{Queries}         & \textbf{First Disagreed Passages}      \\
    \hline
    \multicolumn{2}{l}{\textbf{CAsT Topic-31}} \\ \hline
    What is throat cancer?  & \multirow{9}{0.51\linewidth}{\textbf{\texttt{ConvDR}}: There are two main types of esophageal cancer: squamous cell cancer and adenocarcinoma of the esophagus. Squamous cell cancer occurs most commonly in African Americans as well as people who smoke cigarettes...\\
    \textbf{\texttt{ConvDR-\model}}: \textcolor{red}{In fact, some people diagnosed with throat cancer are diagnosed with esophageal, lung, or bladder cancer} at the same time. This is typically because cancers often have the same risk factors, or because cancer that begins in one part of the body can spread throughout the body... 
    } \\ 
   Is it treatable? &\\
   Tell me about lung cancer.  &\\
    What are its symptoms?  &\\
    Can it spread to the throat? &\\
    \textcolor{blue}{What causes throat cancer?} &\\
    \textcolor{blue}{What is the first sign of it?} &\\
     \underline{\textcolor{blue}{Is it the same as esophageal cancer?}} &\\
    \textbf{Ref}:Is throat cancer the same as esophageal cancer? & \\
    
    \hline
    \multicolumn{2}{l}{\textbf{CAsT Topic-58}} \\ \hline
  \textcolor{blue}{What is a real-time database?}  & \multirow{5}{0.51\linewidth}{{\textbf{\texttt{ConvDR}}}: Examples of what the database describes.\\
    \textbf{\texttt{ConvDR-\model}}: \textcolor{red}{A real-time database} is a database systemwhich uses real-time processing to handle workloads whose state is constantly changing. This differs from traditional databases containing persistent data.
    \textcolor{red}{For example}...
} \\ 
   How does it differ from traditional ones? &\\
    What are the advantages of real-time processing?  &\\
     \underline{What are examples of important \textcolor{blue}{ones}?} &\\
    \textbf{Ref}:What are examples of important \textcolor{blue}{real-time databases}? & \\
     \hline

    \multicolumn{2}{l}{\textbf{CAsT Topic-59}} \\ \hline
Which weekend sports have the most injuries?  & \multirow{9}{0.51\linewidth}{{\textbf{\texttt{ConvDR}}}: To help \textcolor{red}{recover from minor injuries}, overexertion or
surgery, Arnica is a must for every medicine cabinet. Whether you are an active baby boomer... \\
\textbf{\texttt{ConvDR-\model}}: \textcolor{red}{Injury Prevention Basics}. It's always better to \textcolor{red}{prevent an injury} than to recovery from one, so learning and following basic \textcolor{red}{injury} \textcolor{red}{prevention} advice is step one. \textcolor{red}{The best way to avoid injuries} is to be prepared for your sport, both physically and mentally. Don't succumb to the weekend warrior syndrome...
} \\ 
   What are the most common types of injuries? &\\
    What is the ACL?  &\\
    What is an injury for it? &\\
    Tell me about the RICE method. &\\
    Is there disagreement about it? &\\
    What is arnica used for? &\\
     \underline{What are some ways to avoid injury?} &\\
    \textbf{Ref}:What are some ways to \textcolor{blue}{avoid} sports injuries? & \\
     \hline
    \end{tabular}
    }
    \vspace{2mm}
\end{table*}

To verify the effectiveness of the topic segmentation of our method, we conduct an experiment that concatenates different lengths of randomly sampled utterances to the beginning of the current conversation session.
In this experiment, we use the \texttt{ConvDR} as our baseline.
Figure~\ref{fig:add_turn} shows the search performance of our \model and \texttt{ConvDR} with different length of random sampled noise utterances as input.
From Figure~\ref{fig:add_turn}, we find that our \model is more robust to concatenate more random sampled utterances.
When we concatenate more random sampled utterances, the performance of \texttt{ConvDR} dropped dramatically while \texttt{ConvDR-\model} slightly dropped in the beginning and kept stable.
The reason for this phenomenon lies in that our model can identify the topic segmentation boundary and reduce the impact of unrelated utterances when encoding the current conversational query.
This demonstrates that the topic segmentation helps the model focus on the utterances of relevant topics.

\subsection{Case Study}
We show three cases in Table~\ref{tab:case} to intuitively understand how our self-supervised tasks of \model improve the performance of the existing conversational search methods.

In the first case, \texttt{ConvDR}, which equally treats every historical query, struggles with the long dialogue history and retrieves the irrelevant passage. 
After incorporating \model, the topic segmentation makes \texttt{ConvDR-\model} split out several most related utterances in conversational history. 
With the help of modeling the topic boundary, it easily discovers that ``throat cancer'' is the referred term for the current query. 

In the second case, due to the complex historical queries, \texttt{ConvDR} is confused about whether the ``ones'' in the last query means ``database'' or ``real-time database'' and results in a unrelated retrieved passage.
Our proposed coreference identification task makes \texttt{ConvDR-\model} bypass these obstructions and straightly point out the referred query, and \texttt{ConvDR-\model} successfully finds the accuracy result.

The contextual semantic vanishing will harm the performance since the incomplete contextual semantics cannot accurately represent the search intent. 
In the last case, it makes \texttt{ConvDR} misunderstand the meaning of ``avoid'' in the current query to ``recover''. 
Then its retrieved passage mainly illustrates ``how to recover from sports injuries''. 
The word reconstruction demonstrates its effectiveness and keeps the semantic information of ``avoid'', which is indispensable during representation learning. 
The complete contextual semantic leads \texttt{ConvDR-\model} to more accurate retrieval.




\subsection{Parameter Tuning}
\begin{figure}[h!]
\hspace{-5mm}
  \centering
      \includegraphics[width=0.95\linewidth]{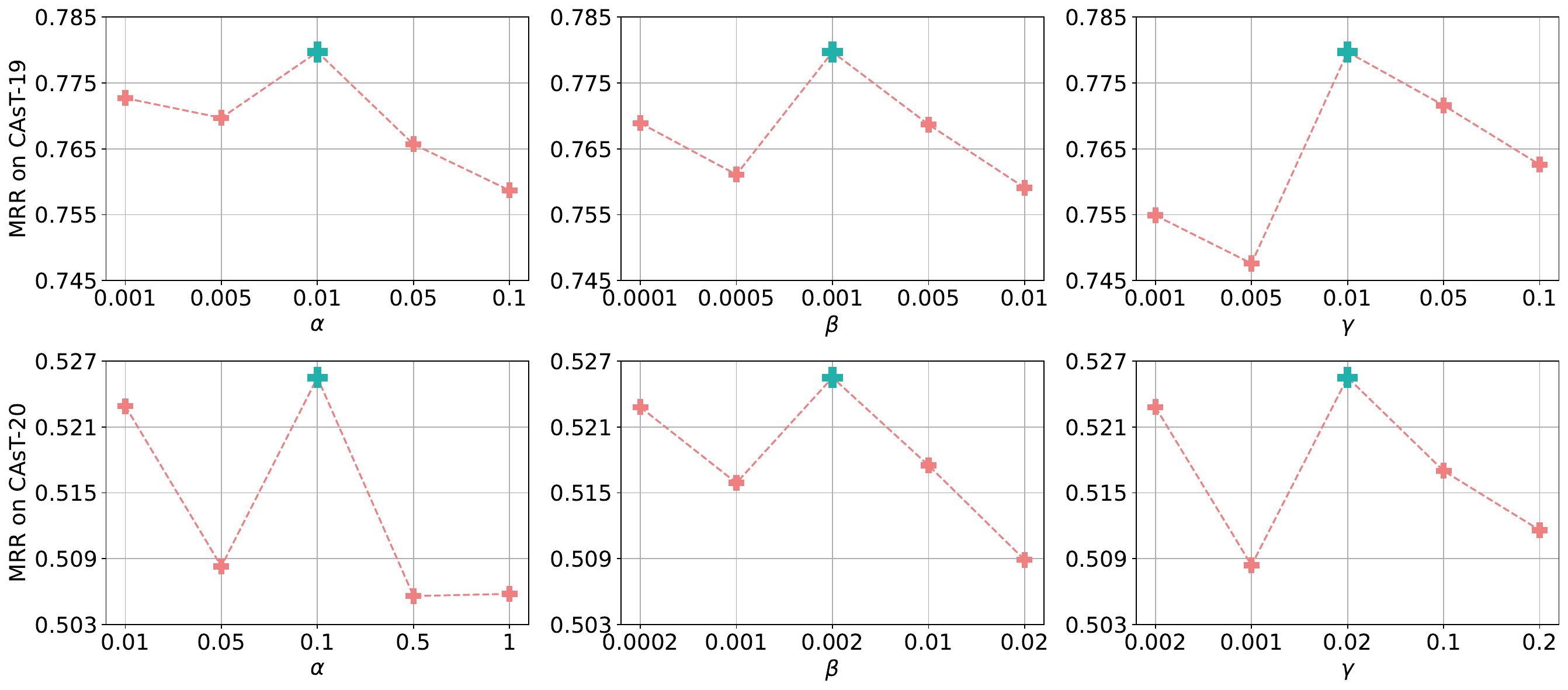}
  \caption{The parameter analysis for weights $\alpha, \beta$ and $\gamma$. }
  \label{param_analysis}
\end{figure}
In this section, we analyze how much the hyper-parameters $\alpha, \beta$, and $\gamma$ influence the retrieval performance and explore the best setting of hyper-parameters.
We design five-group experiments for each parameter and each dataset and the performance comparison as Figure~\ref{param_analysis}.
We find that the performance of \texttt{ConvDR-\model} slightly drops when the parameter changes, and this demonstrates the hyper-parameter robustness of \model. 
Finally, we determine the best setting of $\alpha, \beta$, and $\gamma$ to be $1e-2$, $1e-3$, $1e-2$ for CAsT-19 and $1e-1$, $2e-3$, $2e-2$ for CAsT-20.

\section{Conclusion}
In this work, we propose a novel \fullmodel framework \model for conversational search, which could easily be applied to existing methods and boost their performance.
Different from the conventional warm-up method, our proposed \model introduces three self-supervised tasks to better initialize the conversational encoder. 
These extra supervision signals guide the model to understand complex conversational structure and effectively prevent contextual semantic vanishing.
Extensive experiments conducted on two benchmark datasets prove the effectiveness of \model, which improves previous methods and achieves the best performance.
Extra analytical experiments further answer why our self-supervised tasks could improve performance.

\section{Limitations}
Despite we largely improve the performance of the existing conversational search method, the mechanism of the self-supervised tasks in our \model is simple and intuitive. 
Additionally, our post-training method relies on the external query reformulation dataset, which is a compromise under the scarcity of conversational search data. 
However, the essential contribution of this work is that we point out the significance of modeling dialogue structure (especially for topic shift), and the phenomenon of contextual semantic vanishing in conversational search for the first time.
We hope future works could pay more attention to these problems and devise more complex methods to develop more powerful conversational search systems.

\section*{Acknowledgements}
We would like to thank the anonymous reviewers for their constructive comments. This work was supported by National Natural Science Foundation of China (NSFC Grant No. 62122089 \& No. T2293773), Beijing Outstanding Young Scientist Program NO. BJJWZYJH012019100020098, and Intelligent Social Governance Platform, Major Innovation \& Planning Inter-disciplinary Platform for the ”Double-First Class” Initiative, Renmin University of China. 

\newpage


\bibliography{anthology,custom}
\bibliographystyle{acl_natbib}

\appendix

\newpage
\section{Post-training Dataset}
Followed by the existing conversational dense retrieval methods, we also use the query reformulation dataset for our proposed \model model.
\textbf{QReCC}~\cite{anantha-etal-2021-open} is a query rewriting dataset which contains 14K conversations. 
The queries in QReCC are collected from three sources: TREC CAsT~\cite{dalton2020cast}, QuAC~\cite{choi-etal-2018-quac} and NQ~\cite{kwiatkowski-etal-2019-natural}. 
The queries in NQ were used as prompts to create conversational queries. 

We notice that the queries in TREC CAsT dataset are used in the conversational search fine-tune phrase, it will cause the data leaking problem.
For fair comparison, we filter the queries from TREC CAsT from QReCC. 
The statistics of the filtered QReCC dataset are shown in Table~\ref{tab:statistics of qrecc}. 


\begin{table}[h]
  \centering
  \small
  \caption{The statistics of QReCC after filtering queries from TREC CAsT. (Convs. means conversations, Qrs. means queies, Ref. means reformulation. `\#' indicates count numbers.) }
    \begin{tabular}{l|c|c}
    \toprule
    \textbf{Statistics} & \textbf{$\text{QReCC}_{\text{QuAC}}$}& \textbf{$\text{QReCC}_{\text{NQ}}$} \\
    \midrule
    \# Convs. & 9124  & 4394   \\
    \# Qrs.  & 62749 & 16455  \\ \midrule
    \# Avg. Convs. Tokens & 162.2 & 87.8  \\
    \# Avg. Qrs. Tokens  & 6.5  & 6.7   \\
    \# Avg. Ref. Qrs. Tokens  & 10.4 & 9.1  \\
    \# Avg. Qrs./Convs. & 6.9  & 3.7   \\
    \bottomrule
    \end{tabular}%
  \label{tab:statistics of qrecc}%
  \bigskip
\end{table}%

\label{sec:appendix}


\end{document}